\documentclass{article}

\PassOptionsToPackage{numbers}{natbib}
\usepackage[final]{nips_2017}

\usepackage[utf8]{inputenc} 
\usepackage[T1]{fontenc}    
\usepackage{hyperref}       
\usepackage{url}            
\usepackage{booktabs}       
\usepackage{amsfonts}       
\usepackage{nicefrac}       
\usepackage{microtype}      
\usepackage{graphicx}       
\usepackage{subcaption}     

\title{HELOC Applicant Risk Performance Evaluation by Topological Hierarchical Decomposition}

\author{
  Kyle Brown \\
  Department of Computer Science and Eng. \\
  Wright State University \\
  Dayton, OH 45435 \\
  \texttt{brown.718@wright.edu}
  \And
  Derek Doran \\
  Department of Computer Science and Eng. \\
  Wright State University \\
  Dayton, OH 45435 \\
  \texttt{derek.doran@wright.edu}
  \And
  Ryan Kramer\\
  Air Force Research Laboratory \\
  Wright-Patterson Air Force Base, OH 45433 \\
  \texttt{ryan.kramer.3@us.af.mil} \\
  \And
  Brad Reynolds \\
  Air Force Research Laboratory \\
  Wright-Patterson Air Force Base, OH 45433 \\
  \texttt{reynolds.0345@gmail.com} 
}

\begin{document}
\maketitle

\begin{abstract}
Strong regulations in the financial industry mean that any decisions based
on machine learning need to be explained. This precludes the use of powerful
supervised techniques such as neural networks. In this study we propose a new
unsupervised and semi-supervised technique known as the topological hierarchical
decomposition (THD). This process breaks a dataset down into ever smaller groups,
where groups are associated with a simplicial complex that approximate the underlying
topology of a dataset. We apply THD to the FICO machine learning challenge dataset,
consisting of anonymized home equity loan applications 
using the MAPPER algorithm to build simplicial complexes. We 
identify different groups of individuals
unable to pay back loans, and illustrate how the distribution of feature values in a
simplicial complex can be used to explain the decision to grant or deny a loan by
extracting illustrative explanations from two THDs on the dataset.
\end{abstract}

\section{Introduction}
\label{sec:introduction}
Because of heavy regulations in the financial services industry, there are
stringent requirements for financial decisions made by algorithm to be explainable. 
In particular, it is
important for credit institutions to be able to explain their decisions to
creditors. This 
presents a challenge to the adoption of artificial intelligence techniques
in the industry, as many AI algorithms act as ``black boxes" which are difficult
for a human to interpret or to explain why the algorithm made a certain
decision~\cite{kumar2018opening,dosilovic2018explainable}. Unfortunately, the most
powerful machine learning techniques such as deep neural networks are not
inherently explainable. The goal of explainable AI (XAI) is to remedy this issue
by one of two broad approaches: the first takes existing (unexplainable) algorithms 
to them explainable; the second develops new, powerful explainable techniques
from scratch. This study takes the latter approach. 

We propose a new unsupervised method for decomposing a dataset
into groups sharing similar features, which we call a {\em topological hierarchical
decomposition} (THD). This is done by repeated applications of the MAPPER algorithm
~\cite{singh2007topological} to smaller and smaller subsets of a dataset.
MAPPER estimates the nerve of an open cover constructed on a point-cloud dataset.
It was proposed in the context of topological 
data analysis (TDA), a field which develops
tractable algorithms to compute algebraic structures from topology. Topology is
a branch of mathematics which studies the qualitative structure of spaces, such as
dimension, shape, openness, connectedness, and other complex properties. A good
overview of TDA is given by \citet{carlsson2009topology}. 

We apply THD
to an anonymized dataset of home equity line of credit (HELOC) loan applications made available by 
FICO~\cite{website:fico_heloc} and illustrate how it provides insight, based on groups with
distinct distributions of data features, into why applicants may be unable to pay back a loan within 90 days,
making them risky to loan to.
The goal is not to provide an exhaustive explanation for
every individual from the dataset. Instead, we will provide illustrative explanations extracted
from two THDs and describe how they could be used in an explainable AI approach. Examples include
use by a lending institution to explain to a potential creditor why they are denied or granted
a home equity line of credit, or to explain a decision made by a black-box
supervised machine learning algorithm, simply based on records of past loan performances
rather than by training a transparent supervised learning algorithm that requires 
a historical account of whether customers' past loans were approved or denied. 

\section{Related work}
\label{sec:relwork}
Much work on explainable AI (XAI) has been done in recent years, some
of it concerning financial services. \citet{kumar2018opening} introduce
CLEAR-Trade, which is a system similar to
techniques for visualizing image-classification neural networks, where
instead of finding regions of images that have high activation values
it visualizes time intervals of stock market data
causing intense network activations. 
Relevant to our work here are studies on the prediction of credit risk, with varying
levels of explainability. These include the use of fuzzy support vector
machines~\cite{wang2005new}, non-parametric models~\cite{khandani2010consumer},
neural networks~\cite{khashman2010neural}, and one study which compares
multiple algorithms including neural networks, k-nearest neighbors and
decision trees~\cite{galindo2000credit}. These studies focus on the performance
of the algorithms for classification and regression, rather than the explanation
of the predictions they make.

Much of the practical applications of TDA have been in the biological
sciences and related fields such as chemistry~\cite{offroy2015topological}.
We were unable to find applications of TDA to financial services in the
literature, so the rest of this section will discuss applications of TDA to
other fields.
\citet{nicolau2011topology} applied TDA to a bioinformatic dataset of gene
expression data to identify a new subtype of Estrogen Receptor-positive breast
cancers. This subtype was invisible to traditional cluster analysis, but was
made visible with the application of MAPPER. \citet{offroy2015topological} apply
MAPPER to the Raman spectra of various bacteria and contrast the results with
traditional data analysis techniques. They found that TDA is able to extract
useful information from samples with high signal-to-noise ratio.
\citet{nielson2015topological} use TDA for data-driven discovery in preclinical
traumatic brain injury and spinal cord injury datasets. They mention the possible
utility of TDA in clinical decision making, treatment planning, and rapid
diagnosis.

\section{Methodology}
\label{sec:methodology}
\begin{figure}[t]
	\centering
	\includegraphics[width=0.9\textwidth]{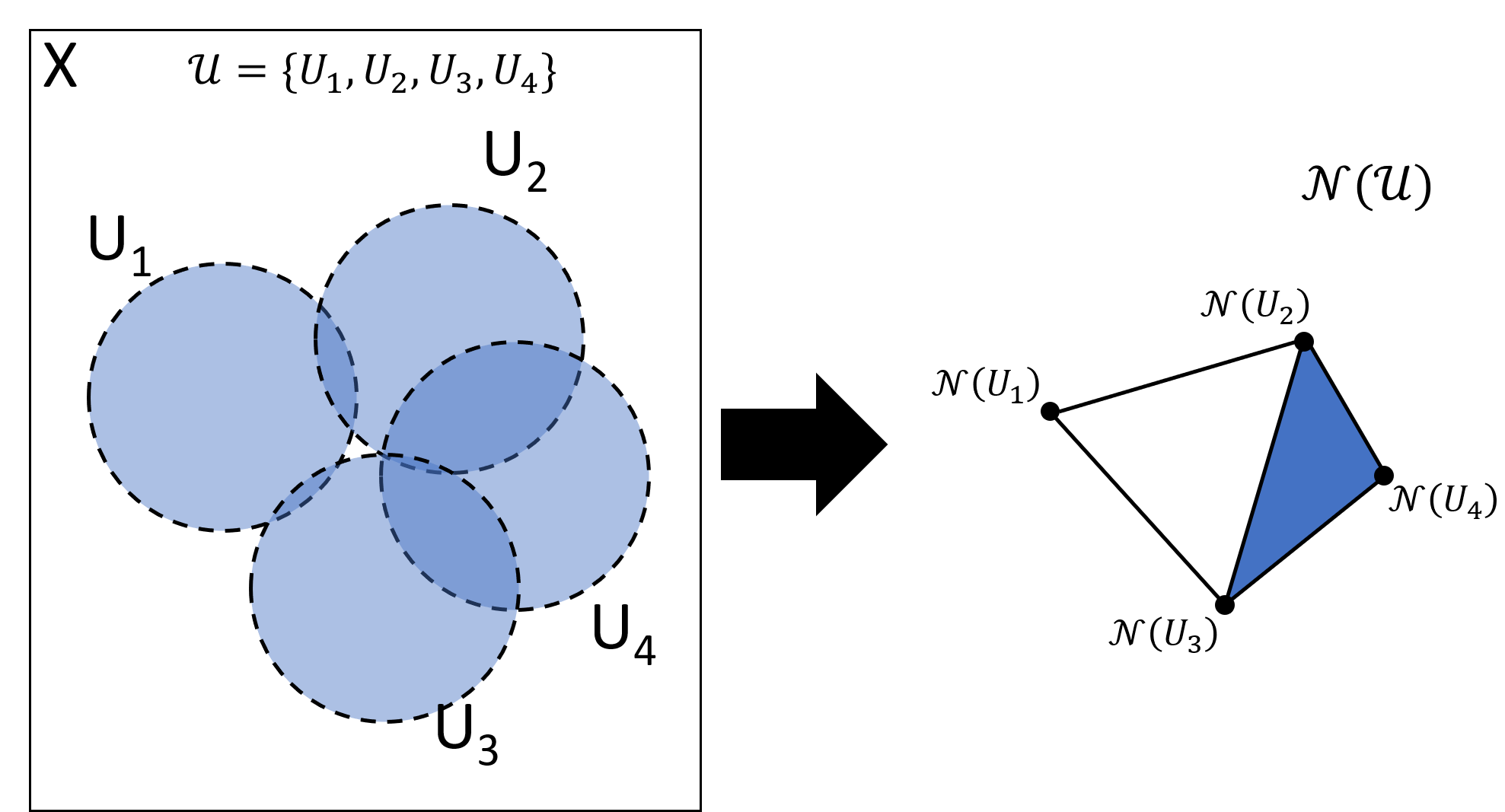}
	\caption{Illustration of the nerve of a simple collection of open sets}
	\label{fig:nerve_diagram}
\end{figure}

In this section we first review the MAPPER construction for approximating 
the nerve of a dataset, and then
describe the topological hierarchical decomposition (THD) in terms of
MAPPER. For background on general topology and metric spaces, consult~\citet{munkres2000topology},
and for algebraic topology and simplicial complexes see~\citet{hatcher2002algebraic}.
Let $X$ and $Z$ be two metric spaces, and $f: X \to Z$
a continuous map between them. We will call $X$ the \emph{data space},
$Z$ the \emph{filter space}, and $f$ a \emph{filter function} or \emph{lens}.
Given an open covering $\mathcal{U} = \left\{U_i\right\}_{i\in I}$
on $Z$, let $f^{*}(\mathcal{U})$ denote the \emph{pullback cover} $\mathcal{V}$
in $X$, which is obtained by splitting the connected components of $f^{-1}(U_i)$
and collecting them to obtain an open cover of $X$. Then the \emph{MAPPER}
construction $\mathcal{M}(X, Z, f, \mathcal{U})$ is the simplicial complex obtained
by taking the nerve of the pullback cover~\cite{singh2007topological}:
\begin{equation}
\mathcal{M}(X, Z, f, \mathcal{U}) = \mathcal{N}\left(f^{*}\left(\mathcal{U}\right)\right)
\end{equation}
An illustration of taking the nerve of a simple open cover is given in 
Figure~\ref{fig:nerve_diagram}. The left panel represents open covers of $X$, each 
defined by $f^{-1}(U_i)$. Open sets become vertices in the resulting simplicial
complex in the right panel and whenever two open sets intersect, there is an edge between their corresponding
vertices in the complex. When three sets intersect, there is a 2-simplex between their
vertices in the complex, represented as a triangle in the figure.

For our purposes, we only need the \emph{1-skeleton} of this simplicial complex,
which can be thought of as an undirected graph which we call a \emph{topological network}.
For an example of such a topological network, see Figure~\ref{fig:thdnetworks}.
In practice, the pullback cover is constructed by binning points in $Z$ and running a clustering
algorithm on each inverse image of a bin in $X$. A common binning technique to generate a cover
is to use generalized intervals, which are Cartesian products of the form 
$\prod_{i=1}^d\left(a_i, b_i\right)$ where $d$ is the dimension of the filter space $Z$. This is
most easily understood in 1-dimension, where typically bins of the form $(a_i - \epsilon, a_i + \epsilon)$
are used, where $\epsilon > 0$ can be viewed as a parameter controlling the size of the bins, 
and therefore their overlap, and $a_i$ are evenly spaced centers of the bins. Alternatively, we can
consider choosing a number of bins $N$ and an overlap parameter $g$ and determine $\epsilon$ from these.
We will denote the resulting open cover on $Z$ by $\mathcal{U}(N, g)$ and call $N$ the \emph{resolution}
of the covering and $g$ the \emph{overlap parameter}, where $0 \le g \le 1$. For more discussion of the mathematics of MAPPER
and its properties, see \citet{dey2016multiscale}.

A \emph{topological hierarchical decomposition} (THD) is an algorithm for decomposing a dataset into smaller
groups based on iterative applications of MAPPER. We start with the entire dataset $X$ and an initial resolution $N_0$,
which is usually very small ($\approx 1$). Keeping the overlap parameter $g$ fixed throughout the entire
process, we run MAPPER on $X$ to obtain the topological network $K$. This network becomes the root node of a tree
structure called the \emph{THD tree}. We then look at the connected components of
this network, say $K_i$. For each $K_i$, if the number of data points $|X_i|$ in this connected component is
above some threshold $t > 0$, we say that we have observed a \emph{split}. For each split, there will be a branch
in the THD tree starting from the current network $K$. We then proceed to recursively split each subset of the
data $X_i$ with a size above the threshold $t$ using the same THD process. If for any network we do not observe
a split, then we increase the initial resolution by some increment $\delta N$ and run MAPPER again on the
corresponding dataset. This creates a linear (i.e. non-branching) path in the THD tree until the next split is observed.
An example of a such a path in a THD tree between splits can be seen in the bottom-center of Figure~\ref{fig:vnenhl_thdtree}.

The idea of a THD should not be confused with the similar idea of \emph{persistent homology}~\cite{zomorodian2005computing},
of which the purpose is to estimate the topological structure of a space from a sample of points in it~\cite{edelsbrunner2008persistent}.
Homology is a \emph{global} structure of a space, determined by the totality of its points. In contrast, THD aims to
\emph{decompose} a space and understand its local features in terms of the resulting groups or clusters. Furthermore,
since at each split the dataset fed to MAPPER changes, there is no persistent structure (such as a persistence module)
defined for THD. For a method of extracting a persistence module from a dataset using MAPPER, consult the 
multi-scale MAPPER of \citet{dey2016multiscale}.

In fact, a THD has more in common with the cluster tree~\cite{stuetzle2003estimating,chaudhuri2010rates},
a hierarchy constructed from a density function that can be estimated with clustering algorithms such 
as single-linkage~\cite{gower1969minimum}. However, unlike these algorithms which fully partition the dataset,
the not all data-points will be present in the leaf nodes of a THD tree. This is similar to density-based
algorithms such as DBSCAN~\cite{ester1996density}, which is able to classify points as \emph{outliers} that
are not included in the high-density clusters~\cite{kriegel2011density}. However, in a THD the outliers are
not based on an (estimated) density but small connected components of a topological graph obtained from
the MAPPER construction.

We implemented the THD process described above to decompose a dataset and examine the distribution of
feature values at each split to determine the significant differences between groups that cause splits. We also
look at the distribution of a label feature in each group which wasn't used in building the THD, and use the significant
features to explain the deviation of this label feature in a group from the global average. From this process we
extract a structure superficially similar to a decision tree, where branches are not determined by the values of a
single feature, but instead by significant differences in the values of a group of features at each split in the THD. Moreover, these splits are constructed in a completely unsupervised way as only the non-label features were used in building the THD, whereas a decision tree is constructed using labels in the
scoring function to decide which feature to split on.
This unsupervised decision-structure may be combined with the decisions made by a supervised learning algorithm to
explain those decisions by tracing the path of a single individual through the THD and looking at significant
characteristics of the groups the individual inhabits.

\section{Results}
\label{sec:results}
This study uses of the FICO Explainable Machine 
Learning dataset (hereafter the \emph{HELOC dataset}) made available by FICO~\cite{website:fico_heloc}.
The dataset consists of anonymized home equity line-of-credit (HELOC) loan applications
made by homeowners requesting a loan in the range of \$5,000 to \$150,000. The target (label) feature
is called RiskPerformance, and is a categorical value of either "Bad" if the consumer was more than
90 days past the due date on a payment in the 24 months after their credit account was opened, and
"Good" otherwise. The dataset contains 5,000 "Good" individuals and 5,459 "Bad" individuals for a total
of 10,459 samples, giving a distribution of 48\% "Good" individuals and 52\% "Bad" individuals.

We did not seek to predict RiskPerformance, but merely explain
its value given the other features in an unsupervised way using a THD. To establish the difference between
groups, we did a statistical comparison between them using KS-score for continuous variables and
a hypergeometric distribution for categorical ones. By doing this for several splits in the THD,
we can then track the path of an individual throughout the THD, using the "choice" of which branch
to follow at each split to "tell a story" about why this person was or was not able to pay back a loan
on time. From these branches we were able to extract illustrative explanations for whether individuals
in a group were 90 or more days delinquent 24 months after taking out a loan.

\begin{figure}[t]
	\centering
	\includegraphics[width=\textwidth]{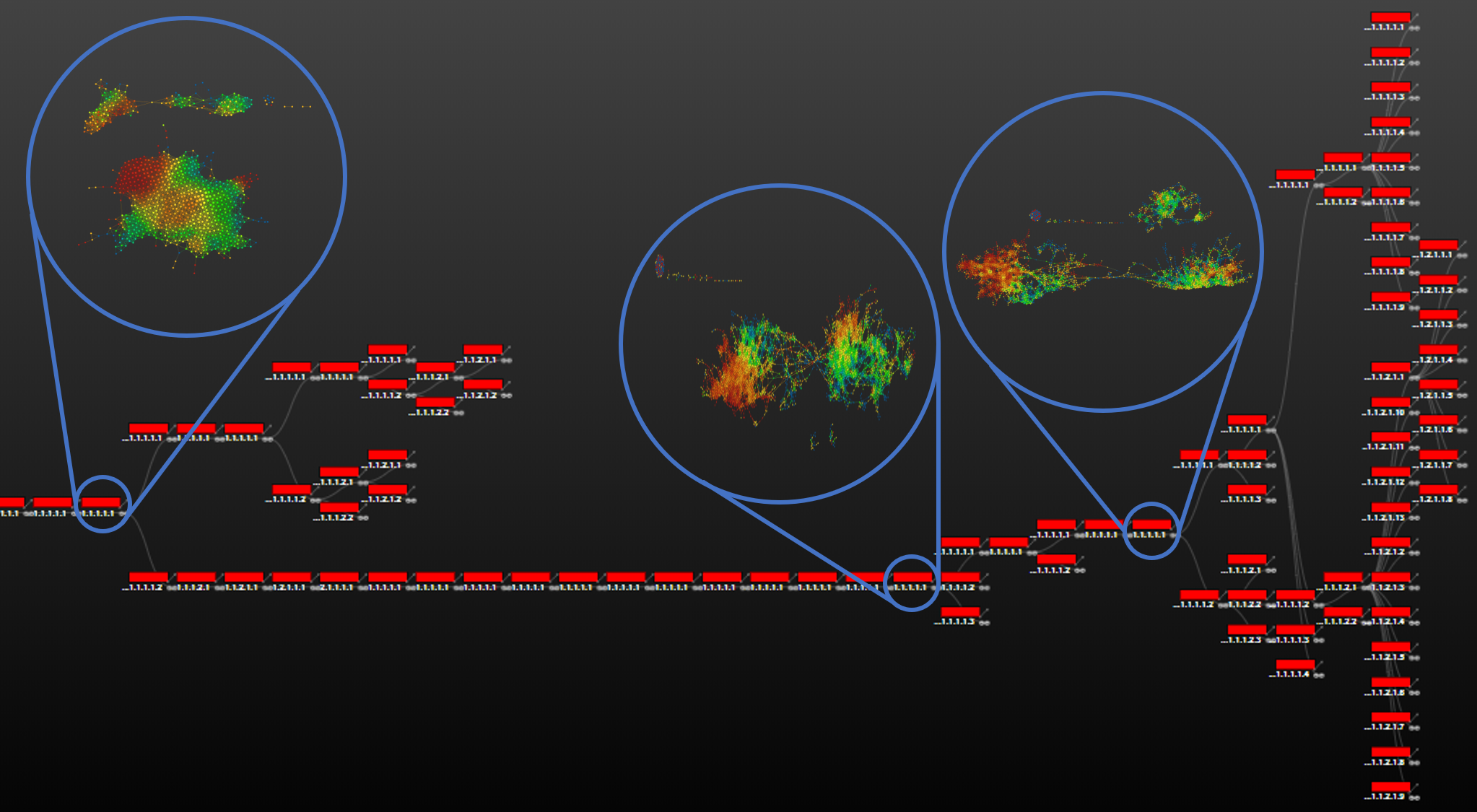}
	\caption{THD tree for VNE metric, NHL as filter with networks for selected groups shown}
	\label{fig:vnenhl_thdtree}
\end{figure}

Two THDs were computed from the entire dataset, using all features but RiskPerformance and ExternalRiskEstimate.
We excluded the ExternalRiskEstimate because it was obtained from an outside source and may not be as
useful in an explanation. For our initial resolution we always used 1, which gives a topological network
with one node containing the entire dataset. This resolution is increased until the first split occurs,
and then THD is ran recursively on each branch until no connected components with a number of points
above the threshold remain. The \emph{gain} (overlap parameter) was 2.7 for all THDs, and remains
fixed throughout the whole process. The split threshold was set to 20 points, i.e. a connected component
with 20 data points (not just nodes) would be considered a split.

We used the variance-normalized Euclidean distance as the metric for both THDs, and
for filters we built one THD with a \emph{neighborhood} lens (NHL), which is analogous to the first two components of t
-SNE~\cite{maaten2008visualizing},
and the other with the first two multi-dimensional scaling (MDS)~\cite{kruskal1964multidimensional,kruskal1964nonmetric}
coordinates. The Ayasdi Platform~\cite{website:ayasdi_platform} was used to build the topological
networks in the THD with MAPPER. The portion of the THD tree for the NHL filter showing splits is given
in Figure~\ref{fig:vnenhl_thdtree}. Topological networks for selected nodes are shown as well, colored by RiskPerformance
where red means more "Good" individuals and blue more "Bad" individuals.
The tree for the MDS filter, not shown, exhibits different behavior in splitting,
where there are a lot of small splits until the last few large splits are reached. In the NHL filter THD,
there is a significant split early on which is not observed with the MDS filter.
In both THDs, there seems to be a large split at the end, with two large groups that are able to be further
decomposed.

\begin{figure}[t]
	\centering
	\includegraphics[width=\textwidth]{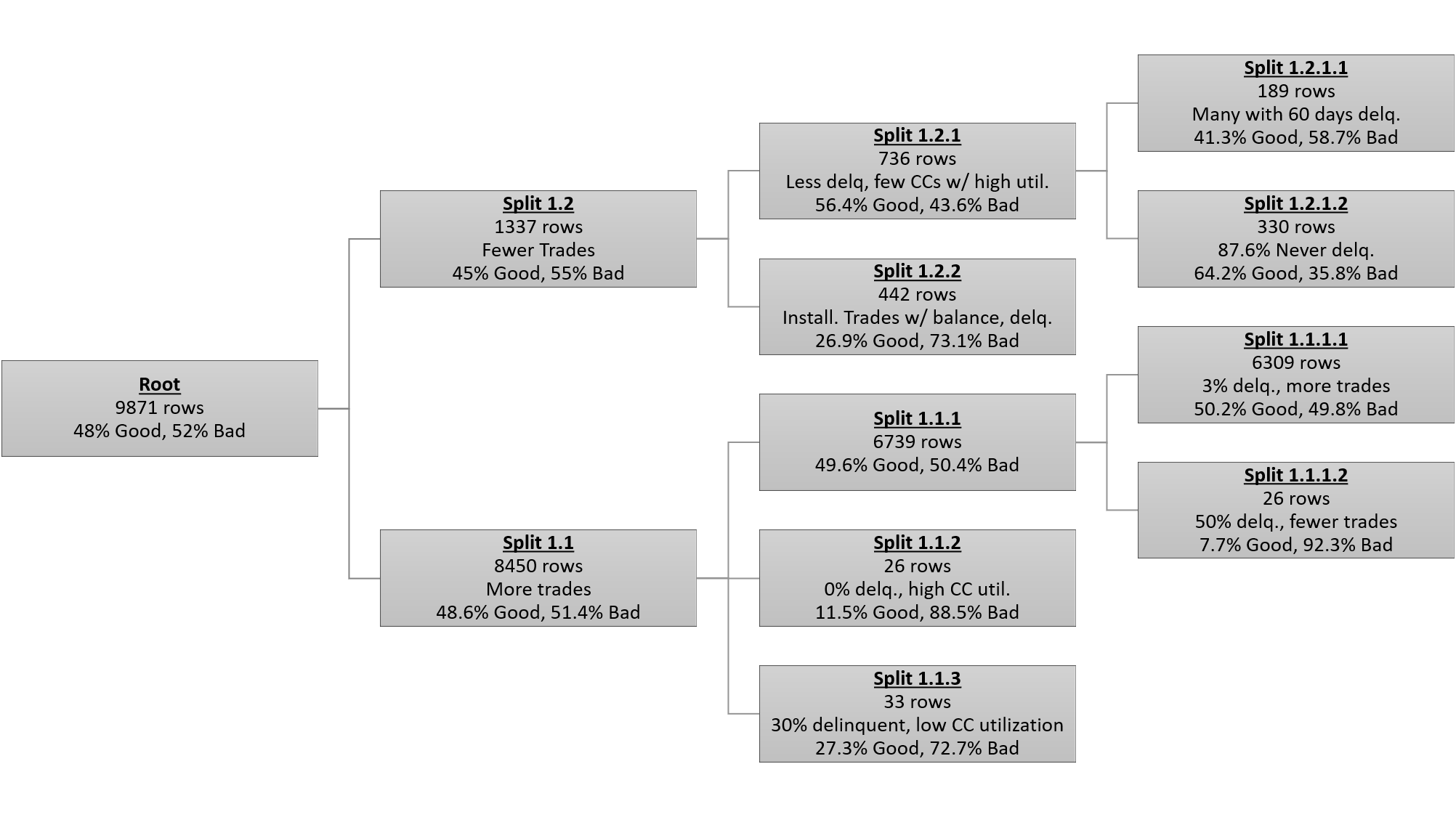}
	\caption{Summary of significant features at high-level splits in the THD tree with VNE metric, NHL as filter}
	\label{fig:vnenhl_summary}
\end{figure}

A simplified view of the NHL THD, showing the early splits, is given in Figure~\ref{fig:vnenhl_summary}. The root node
gives the number of points and RiskPerformance distribution for the entire dataset. At each split, a summary of
the most important features distinguishing this group from other groups in the split is given. Finally, the number
of points and the distribution of RiskPerformance values in the group is given. This diagram could be extended to 
include all splits in a THD, and then used to explain an individual's performance based on their path through the
THD. For example, we observe that individuals falling in Split 1.1.2 could be turned down for a loan as the group has 88.5\% of
its members unable to pay back on time. Further investigation reveals that most individuals in the group exhibit high 
credit card utilization, suggesting an explanation for these individuals. We summarize other explanations are extracted
from Figure~\ref{fig:vnenhl_summary}:
\begin{itemize} 
	\item (a) {Individuals in Split 1.1.2 were unable to pay a loan due to high credit card utilization leading them unable to pay back on time.} 
	\item (b) {Individuals in Split 1.1.3 were unable to pay a loan due to a past history of delinquency, {\em despite} low credit card utilization.}
	\item (c) {Individuals in Split 1.1.1.2 were unable to pay a loan due to having few trades, meaning they have less 
	of a credit history, but also history of delinquency on past trades. This can make such users riskier to lend to.} 
	\item (d) {Individuals in Split 1.2.1.2 paid their loan, even though they have a short history and few trades, but have a very low rate of delinquency.}
\end{itemize}

We extracted explanations in a similar way from the MDS THD as well. Here the group names such as "Split 1.2" refer to groups in the
MDS THD (not shown), and not in the NHL THD:

\begin{itemize}
	\item (e) {Individuals in Split 2 would be denied a loan due to having a large number of loans with balance, and a high number of inquiries.}
	\item (f) {Individuals in Split 1.2 would be denied a loan due to a history of delinquency over 120 days and a large number of trades with balance.}
	\item (g) {Individuals in Split 1.1.1.1.2 would be denied a loan due to a history of delinquency and a high number of revolving trades with balance. This is in spite of the fact that these individuals have a better external risk estimate than 
	individuals in their sister group of the split.}
\end{itemize}

Note in explanation (d) how the THD is able to identify fine grained groups of customers who could be seen as good
loan customers, even though they have a shallow credit history. The THD is further able to find a set of customers
likely to be poor loan customers by explanation (b), 
even though they have relatively low credit utilization. These customers may be 
counterintuitive
in nature, in the sense that their features intuitively suggest that the customer should (not) be granted credit. 

Predicting the group of a split where a new applicant would fall within the THD would thus provide both a decision and reason for
granting or denying the applicant even when applicant features take on surprising or contradictory 
values, and the explanation can be presented at a finer or coarser grain depending on the 
split depth an analyst chooses to select an explanation from. Specifically, 
the denial of a loan can be explained by an applicant that falls into a group with a high percentage 
(greater than the global average of 52\%) of "Bad"
RiskPerformance values, where membership in a group is defined by a distribution of feature values that distinguish the group from
others at a split in the THD. 
These explanations are only based on the features
of the applicant, and have nothing to do with the past loaning behavior of the organization. 
Note that using different settings for the THD will result in different explanations, although there
are some similarities such as a history of delinquency and a large number of loans with balance correlating with "Bad" 
RiskPerformance, and hence these individuals would be denied a loan.

\begin{figure}[t]
	\centering
	\begin{subfigure}{0.48\textwidth}
		\centering
		\includegraphics[width=\textwidth]{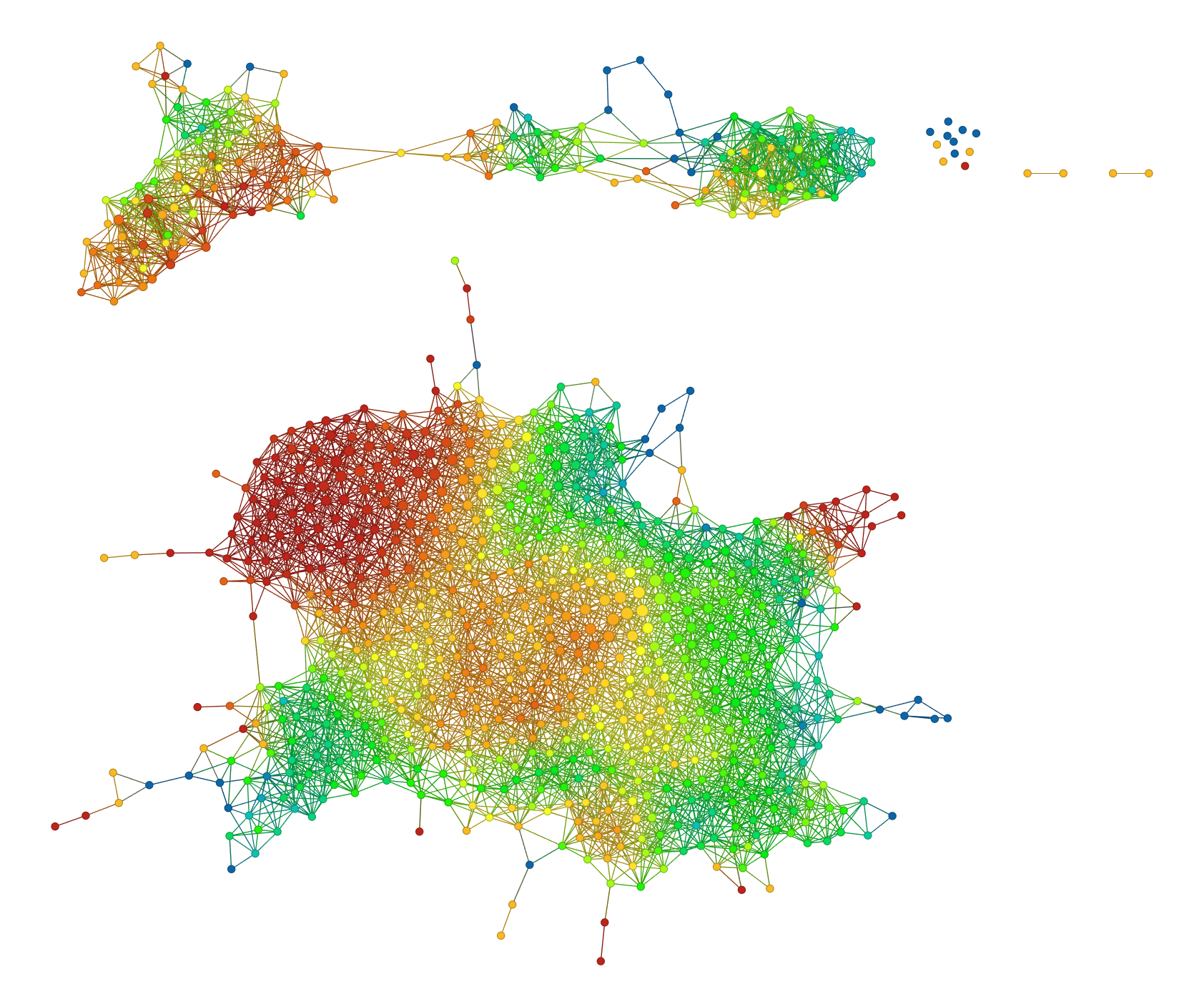}
		\caption{RiskPerformance \\ (blue=bad, red=good)}
	\end{subfigure}
	\begin{subfigure}{0.48\textwidth}
		\centering
		\includegraphics[width=\textwidth]{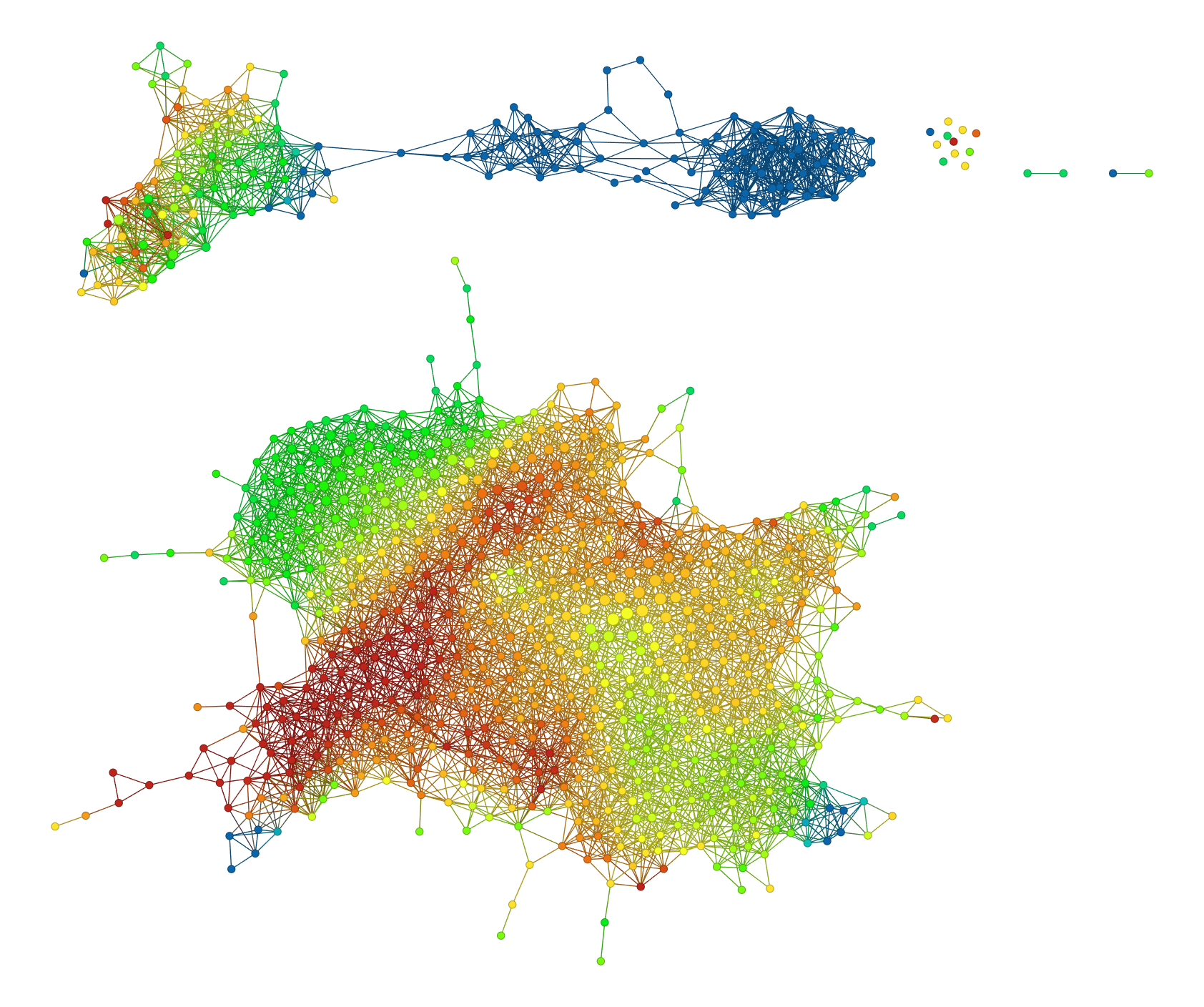}
		\caption{credit cards with high utilization \\ (blue=less, red=more)}
		\label{fig:vnenhl_ntwk_ccwutil}
	\end{subfigure}
	\begin{subfigure}{0.48\textwidth}
		\centering
		\includegraphics[width=\textwidth]{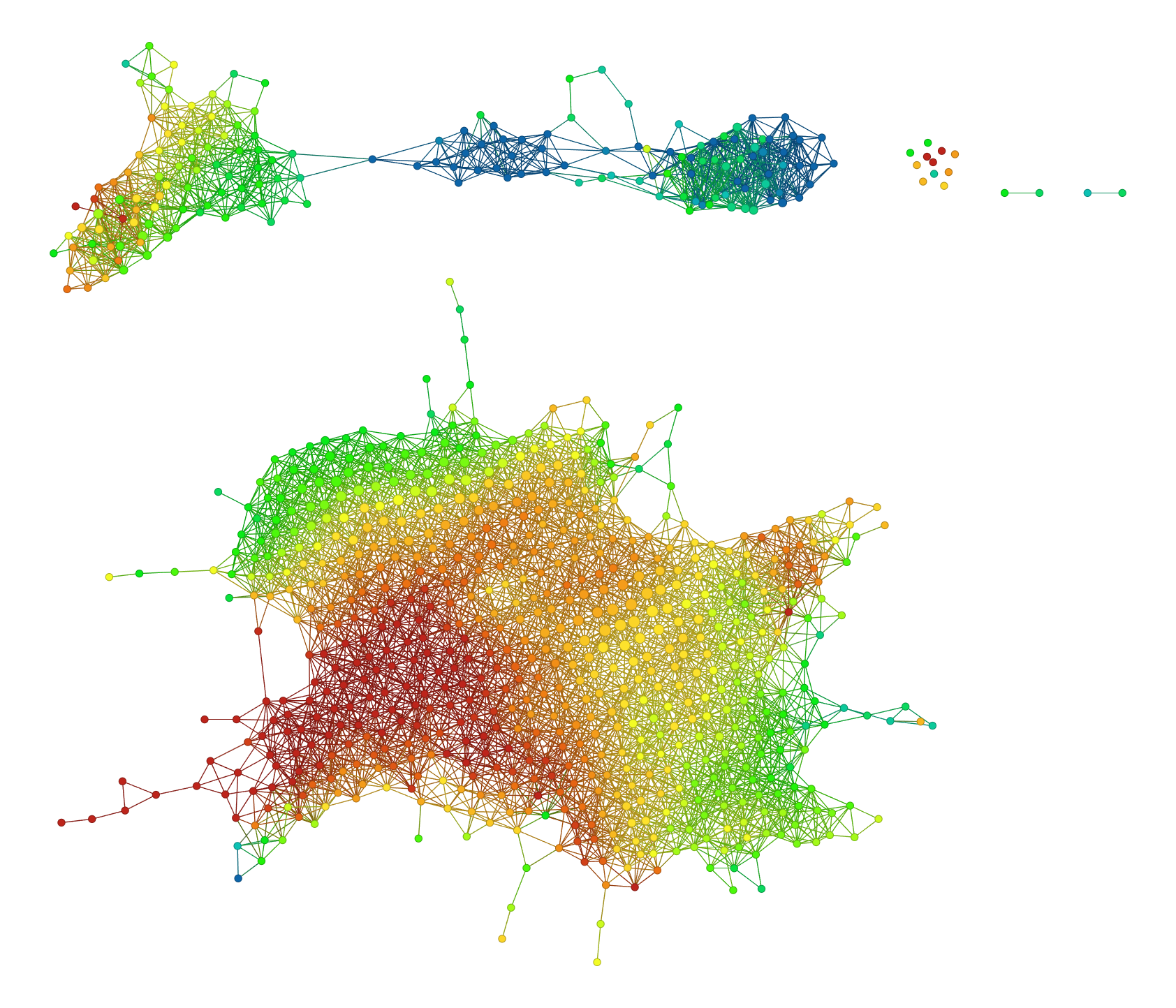}
		\caption{revolving trades with balance \\ (blue=less, red=more)}
	\end{subfigure}
	\begin{subfigure}{0.48\textwidth}
		\centering
		\includegraphics[width=\textwidth]{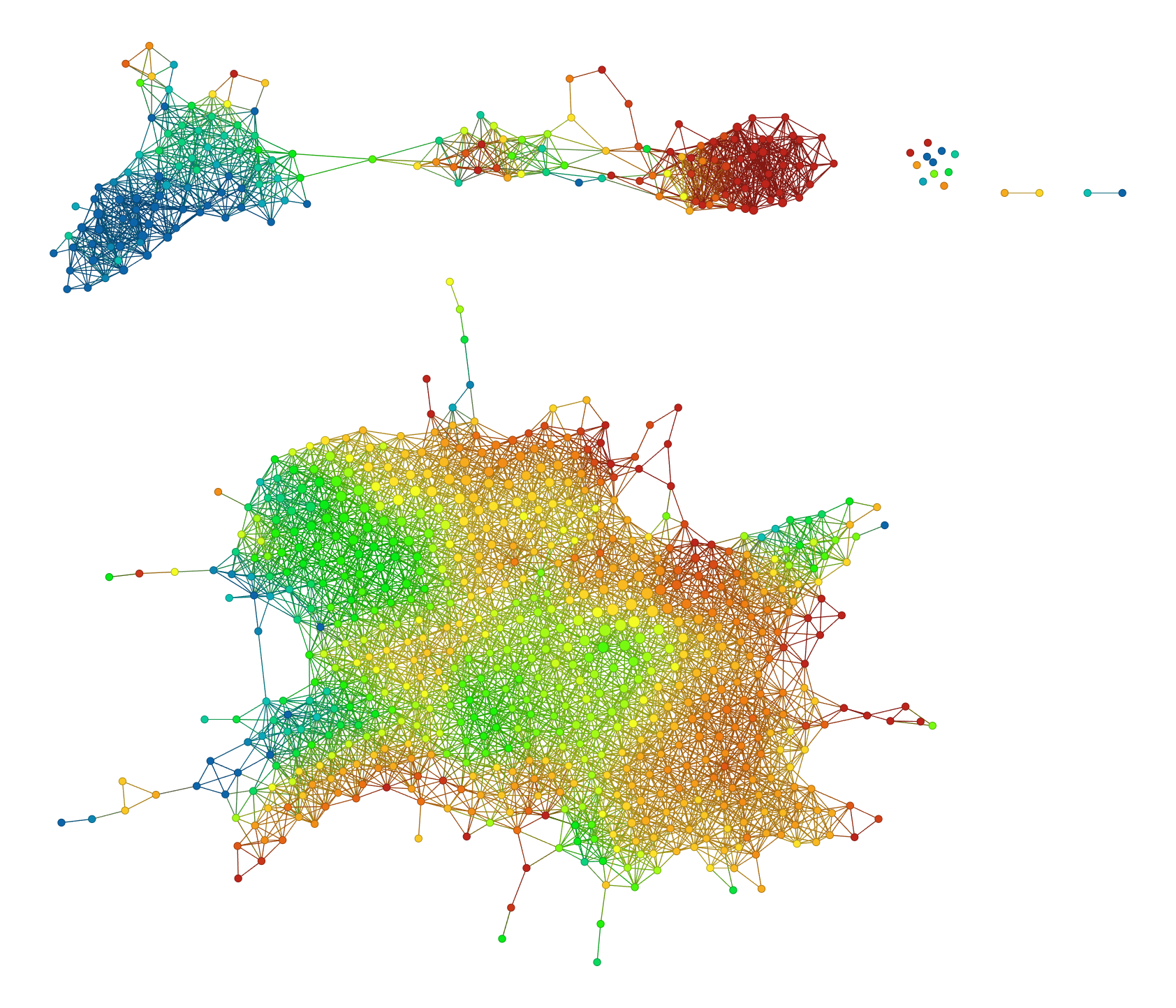}
		\caption{percentage installment trades \\ (blue=0\%, red=100\%)}
	\end{subfigure}
	\caption{Topological Network for the first split in the NHL THD, colored by different features}
	\label{fig:thdnetworks}
\end{figure}

It is also instructive to look at individual topological networks where a split occurs. An example for the first split
in the NHL THD is given in Figure~\ref{fig:thdnetworks}. The connected component at the top of the network corresponds
to the smaller group labeled "Split 1.2" in Figure~\ref{fig:vnenhl_summary}. The further split in this group can already
be seen, as there is a vertex in the upper component that can be removed to split it into two more components. "Split 1.2.1,"
which has few credit cards with high utilization, can be seen as the right component of this upper network, based on the
coloring in Figure~\ref{fig:vnenhl_ntwk_ccwutil}. The topological networks can be used for an even more granular explanation,
as we can consider the nodes an individual belongs to in the network as containing similar individuals. We can also look
at the immediate neighbors of these nodes to get a very local group of individuals similar to the one under consideration.
This ability to go from a high-level, group-based representation, to individual topological networks, and then to just points
from a group of related nodes in a network is a novel and useful feature of THD.

\subsection*{Comparison to transparent supervised models}
It should be noted that Figure~\ref{fig:vnenhl_summary} does not appear all that different from a decision tree.
Each split in the THD is based on a set of feature properties that differentiate one group from another, 
which is not unlike a decision tree that makes classification decisions by learning a hierarchy of 
heuristics to bin data. Moreover, decision trees are inherently transparent in the sence that each path 
down a tree from root to leaf describes a series of conditions explaining why data is classified. 

The key difference between using splits of a THD to provide explanations rather than a decision tree is 
that THD is an entirely {\em unsupervised} technique; in constructing a THD the target feature RiskPerformance
is never used. A decision tree, in contrast, is a {\em supervised} approach where the target feature is used 
directly during learning. When this training data is collected based on credit award decisions made by an organization
from the past, the decision tree essentially learns a model describing how and why a firm awards credit to applicants.
The learned model thus incorporates any 
potentially historical biases or priorities of the organization the training data is from. In taking an unsupervised
approach, the THD becomes {\em decoupled} from the organization or institution who issues credit: splits in a THD are based on 
distinguishing features between sets of past applications conditioned on whether they successfully paid their loan. Thus, 
the THD can lead to automatic loan decisions based solely on the merits of the applicant, instead of a combination of 
applicant merit and historical firm behavior.
Moreover, a THD is theoretically grounded by exploiting the shape and structure of the underlying manifold of 
data about applicants, which is more likely to have a shape and structure characteristics 
across applicants for all forms of credit besides HELOC. Insights from a THD are thus more likely to be 
transferable across domains (e.g., to support decisions for other lines of credit besides HELOC), 
compared to decision tree heuristics that are (over)fitted to a single, specific dataset. 

We further note that the THD requires no \emph{a priori} information about the meaning or importance
of each feature. Since these explanations are independent of any machine learning model used in classification,
they could thus be used to supplement and explain decisions made by the algorithm. For example, 
a linear regression may give larger
magnitude to weights that were found to correlate with RiskPerformance in THD groups, such as percentage of trades never
delinquent and number of trades with balance. Finally, these explanations could also be used to understand a
\emph{misclassification} made by a classifier. The classifier may be weighting the wrong features, i.e. features that correlate
with RiskPerformance in a different THD group than the one the point being classified belongs to. Another possibility is
that the data point being classified is an outlier - it is in a THD group but has unusual features for that group. THD provides
a framework for identifying such points automatically.

\section{Conclusion}
\label{sec:conclusion}
In this paper we introduced the \emph{topological hierarchical decomposition}, 
which constructs a tree-like structure of
topological networks by iteratively applying the MAPPER
construction to smaller subsets of a dataset, based on connected components
of previously computed networks. We described THD based on the parametrization
of a family of open covers on the filter space in terms of a resolution parameter,
and contrasted it with the ideas of persistent homology and hierarchical clustering. 
We then constructed two THDs
on the FICO Machine Learning Challenge dataset consisting of anonymized HELOC
applications. We showed how these THDs can be used to explain related groups of 
features in the dataset and their influence on the target RiskPerformance feature.
We showed examples of THD trees and topological networks that can be used to understand
a dataset, and how colorings of a topological network can be used to understand splits
in a THD.

An important topic that future studies on THD should cover is an understanding
of its mathematical and algorithmic properties. In order to tie THD more closely to
persistent homology, it may become necessary to apply some notion of 
\emph{zigzag persistence}~\cite{carlsson2009zigzag}, which allows one to "study the
persistence of topological features across a family of spaces or point-cloud sets"~\cite{Carlsson2010}.
The "sequence of topological spaces" would be the groups obtained from a THD and the functions
between them just inclusion from the smaller group to its parent, since the smaller group is
always a subset of the parent group. This describes a hierarchical structure of inclusions,
proceeding from leaf nodes in the THD back up to the group containing the entire dataset.
Whether one actually obtains a persistent structure from this remains to be seen.

Other future work could look at applications of THD. One could construct a classifier from a
THD by constructing one from both training and testing data and then looking at the leaf-level
network a testing point falls in and doing a majority vote on the label of its nearest training
points based on nodes in the network. This classifier could then be compared with the state-of-the-art,
and if competitive, could provide a more explainable alternative to them. The amount of hyper-parameters
that need to be selected by a human to construct a THD is large. It would be useful to have some way
to automatically select these parameters to obtain the "optimal" THD. One approach in a classification
context would be to construct several THDs on the training set and then to select the one with the
largest information gain based on the label.

Topological hierarchical decomposition is a versatile tool that can be used for an unsupervised or
semi-supervised approach to data analysis. It also provides a new method for understanding predictions
made by existing supervised algorithms. Networks in a THD tree can be queried at multiple levels
of granularity, providing corresponding levels of explainability. Since it makes use of topological
data analysis, it comes with built-in robustness to noisy data. These features make it useful in almost
any field where analysis of big data is required.

\subsubsection*{Acknowledgments}
This work was supported by the Air Force Research Laboratory, 711th Human Performance Wing, Airman Systems Directorate with funding provided through Oak Ridge Institute for Science and Education (ORISE). Our work has also been supported by the Ohio Federal Research Network project \textit{Human-Centered Big Data}. Any opinions, findings, and conclusions or recommendations expressed in this article are those of the author(s) and do not necessarily reflect the views of the Ohio Federal Research Network. 

The authors would also like to thank Matthew Piekenbrock for discussions on multiscale
MAPPER and hierarchical clustering that were useful in preparing the discussion of
THD and comparisons with other techniques in Section~\ref{sec:methodology}.

\bibliographystyle{plainnat}
\bibliography{feap2018} 

\end{document}